\documentclass[10pt,twocolumn,letterpaper]{article}

\usepackage{cvpr}              

\usepackage{graphicx}
\usepackage{amsmath}
\usepackage{amssymb}
\usepackage{booktabs}
\usepackage{xcolor}
\usepackage{enumitem}
\usepackage{amsfonts}

\usepackage[accsupp]{axessibility}

\usepackage[pagebackref,breaklinks,colorlinks]{hyperref}

\usepackage[capitalize]{cleveref}
\crefname{section}{Sec.}{Secs.}
\Crefname{section}{Section}{Sections}
\Crefname{table}{Table}{Tables}
\crefname{table}{Tab.}{Tabs.}

\def\cvprPaperID{4326} 
\def\confName{CVPR}
\def\confYear{2022}

\begin{document}

\title{Learning To Recognize Procedural Activities with Distant Supervision} 


\author{Xudong Lin$^{1}$\thanks{Research done while XL was an intern at Facebook AI Research.} \quad
Fabio Petroni$^2$  \quad
Gedas Bertasius$^3$ \quad
\\
Marcus Rohrbach$^2$ \quad
Shih-Fu Chang$^{1}$ \quad
Lorenzo Torresani$^{2,4}$  
\\
 $^{1}$Columbia University \quad $^{2}$Facebook AI Research \quad $^{3}$UNC Chapel Hill  \quad $^{4}$Dartmouth \quad
\\
}
\maketitle

\begin{abstract}
In this paper we consider the problem of classifying fine-grained, multi-step activities (e.g., cooking different recipes, making disparate home improvements, creating various forms of arts and crafts) from long videos spanning up to several minutes. Accurately categorizing these activities requires not only recognizing the individual steps that compose the task but also capturing their temporal dependencies. This problem is dramatically different from traditional action classification, where models are typically optimized on videos that span only a few seconds and that are manually trimmed to contain simple atomic actions. While step annotations could enable the training of models to recognize the individual steps of procedural activities, existing large-scale datasets in this area do not include such segment labels due to the prohibitive cost of manually annotating temporal boundaries in long videos. To address this issue, we propose to automatically identify steps in instructional videos by leveraging the distant supervision of a textual knowledge base (wikiHow) that includes detailed descriptions of the steps needed for the execution of a wide variety of complex activities. Our method uses a language model to match noisy, automatically-transcribed speech from the video to step descriptions in the knowledge base. We demonstrate that video models trained to recognize these automatically-labeled steps (without manual supervision) yield a representation that achieves superior generalization performance on four downstream tasks: recognition of procedural activities, step classification, step forecasting and egocentric video classification.
\end{abstract}

\section{Introduction}
Imagine being in your kitchen, engaged in the preparation of a sophisticated dish that involves a sequence of complex steps. Fortunately, your J.A.R.V.I.S.\footnote{A fictional AI assistant in the Marvel Cinematic Universe.} comes to your rescue. It actively recognizes the task that you are trying to accomplish and guides you step-by-step in the successful execution of the recipe. The dramatic progress witnessed in activity recognition~\cite{TSN,Kinetics,tran2018closer,Damen2018EPICKITCHENS} over the last few years has certainly made these fictional scenarios a bit closer to reality. Yet, it is clear that in order to attain these goals we must extend existing systems beyond atomic-action classification in trimmed clips to tackle the more challenging problem of understanding procedural activities in long videos spanning several minutes. Furthermore, in order to classify the procedural activity, the system must not only recognize the individual semantic steps in the long video but also model their temporal relations, since many complex activities share several steps but may differ in the order in which these steps appear or are interleaved. For example, ``beating eggs'' is a common step in many recipes which, however, are likely to differ in the preceding and subsequent steps. 

In recent years, the research community has engaged in the creation of several manually-annotated video datasets for the recognition of procedural, multi-step activities. However, in order to make detailed manual annotations possible at the level of both segments (step labels) and videos (task labels), these datasets have been constrained to have a narrow scope or a relatively small scale. Examples include video benchmarks that focus on specific domains, such as recipe preparation or kitchen activities~\cite{breakfast, rohrbach2012database, youcook2, Damen2018EPICKITCHENS}, as well as collections of instructional videos manually-labeled for step and task recognition~\cite{zhukov2019cross, tang2020comprehensive}. Due to the large cost of manually annotating temporal boundaries, these datasets have been limited to a small size both in terms of number of tasks (about a few hundreds activities at most) as well as amount of video examples (about 10K samples, for roughly 400 hours of video). While these benchmarks have driven early progress in this field, their limited size and narrow scope prevent the training of modern large-capacity video models for recognition of general procedural activities. 

On the other end of the scale/scope spectrum, the HowTo100M dataset~\cite{miech2019howto100m} stands out as an exceptional resource. It is over 3 orders of magnitude bigger than prior benchmarks in this area along several dimensions: it includes over 100M clips showing humans performing and narrating more than 23,000 complex tasks for a total duration of 134K hours of video.  The downside of this massive amount of data is that its scale effectively prevents manual annotation. In fact, all videos in HowTo100M are unverified by human annotators. 
While this benchmark clearly fulfills the size and scope requirements needed to train large-capacity video models, its lack of segment annotations and the unvalidated nature of the videos impedes the training of accurate step or task classifiers.

In this paper we present a novel approach for training models to recognize procedural steps in instructional video {\em without} any form of manual annotation, thus enabling optimization on large-scale unlabeled datasets, such as HowTo100M. We propose a {\em distant supervision} 
framework that leverages a textual knowledge base as a guidance to automatically identify segments corresponding to different procedural steps in video. Distant supervision has been used in Natural Language Processing~\cite{snow2005, mintz2009distant, riedel2010modeling} to mine relational examples from noisy text corpora using a knowledge base. In our setting, we are also aiming at relation extraction, albeit in the specific setting of identifying video segments relating to semantic steps. The knowledge base that we use is wikiHow~\cite{wikihow}---a crowdsourced multimedia repository containing over 230,000 ``how-to'' articles describing and illustrating steps, tips, warnings  and requirements to accomplish a wide variety of tasks. Our system uses language models to compare segments of narration automatically transcribed from the videos to the textual descriptions of steps in wikiHow. The matched step descriptions serve as distant supervision to train a video understanding model to learn step-level representations. Thus, our system uses the knowledge base to mine step examples from the noisy, large-scale unlabeled video dataset. To the best of our knowledge, this is the first attempt at learning a step video representation with distant supervision.

We demonstrate that video models trained to recognize these pseudo-labeled steps in a massive corpus of instructional videos, provide a general video representation transferring effectively to four different downstream tasks on new datasets. Specifically, we show that we can apply our model to represent a long video as a sequence of step embeddings extracted from the individual segments. Then, a shallow sequence model (a single Transformer layer~\cite{vaswani2017attention}) is trained on top of this sequence of embeddings to perform temporal reasoning over the step embeddings. Our experiments show that such an approach yields state-of-the-art results for classification of procedural tasks on the labeled COIN dataset, outperforming the best reported numbers in the literature by more than $16\%$. Furthermore, we use this testbed to make additional insightful observations: 

\begin{enumerate}[noitemsep,nolistsep]
\item Step labels assigned with our distant supervision framework yield better downstream results than those obtained by using the unverified task labels of HowTo100M.  
\item Our distantly-supervised video representation outperforms {\em fully-supervised} video features trained with action labels on the large-scale Kinetics-400 dataset~\cite{Kinetics}.
\item Our step assignment procedure produces better downstream results than a representation learned by directly matching video to the ASR narration~\cite{miech2020end}, thus showing the value of the distant supervision framework.
\end{enumerate}

We also evaluate the performance of our system for classification of procedural activities on the Breakfast dataset~\cite{breakfast}. Furthermore, we present transfer learning results on three additional downstream tasks on datasets different from that used to learn our representation (HowTo100M): step classification and step forecasting on COIN, as well as categorization of egocentric videos on EPIC-KITCHENS-100~\cite{damen2020epic}. On all of these tasks, our distantly-supervised representation achieves higher accuracy than previous works, as well as additional baselines that we implement based on training with full supervision. These results provide further evidence of the generality and effectiveness of our unsupervised representation for understanding complex procedural activities in videos. We will release the code and the automatic annotations provided by our distant supervision\footnote{Please check \url{https://github.com/facebookresearch/video-distant-supervision} for updates.}. 

\begin{figure*}[t]
\begin{center}
\includegraphics[width=0.95\linewidth]{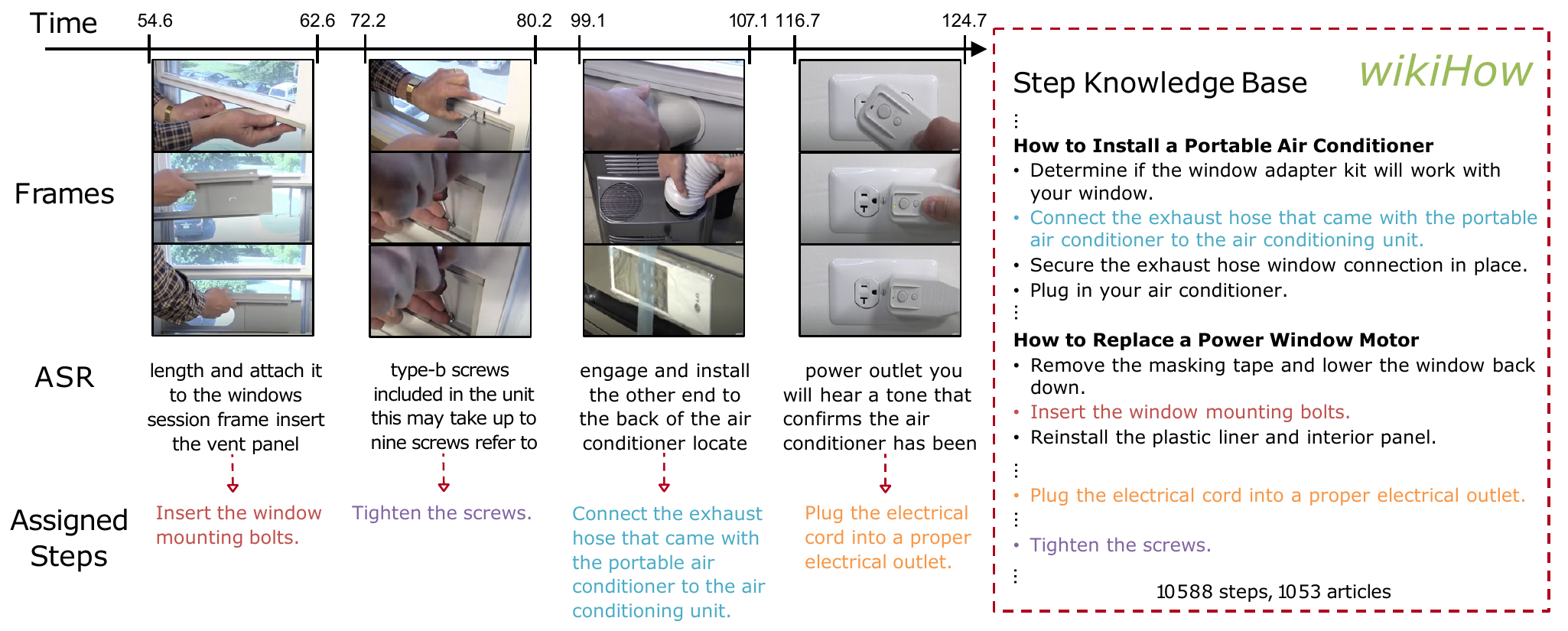}\vspace{-.4cm}

\end{center}
\caption{Illustration of our proposed framework. Given a long instructional video as input, our method generates distant supervision by matching segments in the video to steps described in a knowledge base (wikiHow). The matching is done by comparing the automatically-transcribed narration to step descriptions using a pretrained language model. This distant supervision is then used to learn a video representation recognizing these automatically annotated steps. This video is from the HowTo100M dataset. More examples are provided in the supplementary material. 
}
\label{illustration}
\end{figure*}

\section{Related Work}

During the past decade, we have witnessed dramatic progress in action recognition. However, the benchmarks in this field consist of brief videos (usually, a few seconds long) trimmed to contain the individual atomic action to recognize~\cite{kuehne2011hmdb, UCF101, kay2017kinetics, goyal2017something}. In this work, we consider the more realistic setting where videos are untrimmed, last several minutes, and contain sequences of steps defining the complex procedural activities to recognize (e.g., a specific recipe, or a particular home improvement task).

\noindent\textbf{Understanding Procedural Videos.} Procedural knowledge is an important part of human knowledge~\cite{anderson1982acquisition,rasmussen1983skills,tan2021comprehensive} essentially answering ``how-to'' questions.
Such knowledge is displayed in long procedural videos ~\cite{rohrbach15ijcv,breakfast, youcook2, Damen2018EPICKITCHENS,zhukov2019cross, tang2020comprehensive,miech2019howto100m}, which have attracted active research in recognition of multi-step activities~\cite{hussein2019timeception,hussein2020timegate,zhou2021graph}. Early benchmarks in this field contained manual annotations of steps within the videos~\cite{youcook2, zhukov2019cross, tang2020comprehensive} but were relatively small in scope and size. The HowTo100M dataset~\cite{miech2019howto100m}, on the other hand, does not contain any manual annotations but it is several orders of magnitude bigger and the scope of its ``how-to'' videos is very broad. An instructional or how-to video contains a human subject demonstrating and narrating how to accomplish a certain task. Early works on HowTo100M have focused on leveraging this large collection for learning models that can be transferred to other tasks, such as action recognition~\cite{miech2019howto100m,miech2020end,alayrac2020self}, video captioning~\cite{youcook2,luo2020univl,huang2020multimodal}, or text-video retrieval~\cite{xu2021vlm,miech2020end,bain2021frozen}. The problem of recognizing the task performed in the instructional video has been considered by Bertasius \etal \cite{bertasius2021space}. However, their proposed approach does not model the procedural nature of instructional videos.\\
\noindent\textbf{Learning Video Representations with Limited Supervision.}
Learning semantic video representations~\cite{srivastava2015unsupervised,qiu2017learning,VideoBERT,li2020hero,xu2015discriminative} is a fundamental problem in video understanding research. The representations pretrained from labeled datasets are limited by the pretraining domain and the predefined ontology. Therefore, many attempts have been made to obtain video representations with less human supervision. In the unsupervised setting, supervision signal is usually constructed by augmenting videos~\cite{srivastava2015unsupervised,wei2018learning,feichtenhofer2021large}. For example, Wei \etal \cite{wei2018learning} proposed to predict the order of videos as the supervision to learn order-aware video representations. In the weakly supervised setting, the supervision signals are usually obtained from hashtags~\cite{ghadiyaram2019large}, ASR transcriptions~\cite{miech2019howto100m}, or meta-information extracted from the Web~\cite{gan2016webly}. Miech \etal~\cite{miech2019howto100m} show that ASR sentences extracted from audio can serve as a valuable information source to learn video representations. Previous works~\cite{elhamifar2019unsupervised,elhamifar2020self} have also studied learning to localize keyframes using task labels as supervision. This is different from the focus of this paper, which addresses the problem of learning step-level representations from unlabeled instructional videos.\\
\noindent\textbf{Distant Supervision.}
Distant supervision \cite{mintz2009distant,zeng2015distant} has been studied in natural language processing and generally refers to a training scheme where supervision is obtained by automatically mining examples from a large noisy corpus utilizing a clean and informative knowledge base. It has been shown to be very successful on the problem of relation extraction. For example, Mintz \etal~\cite{mintz2009distant} leverage knowledge from Freebase \cite{bollacker2008freebase} to obtain supervision for relation extraction. However, the concept of distant supervision has not been exploited in video understanding. Huang \etal \cite{huang2020multimodal} have proposed to use wikiHow as a textual dataset to pretrain a video captioning model but the knowledge base is not used to supervise video understanding models.

\section{Technical Approach}
Our goal is to learn a segment-level representation to express a long procedural video as a sequence of step embeddings. 
The application of a sequence model, such as a Transformer, on this video representation can then be used to perform temporal reasoning over the individual steps. Most importantly, we want to learn the step-level representation without manual annotations, so as to enable training on large-scale unlabeled data. The key insight leveraged by our framework is that knowledge bases, such as wikiHow, provide detailed textual descriptions of the steps for a wide range of tasks. In this section, we will first describe how to obtain distant supervision from wikiHow, then discuss how the distant supervision can be used for step-level representation learning, and finally, we will introduce how our step-level representation is leveraged to solve several  downstream problems.

\subsection{Extracting Distant Supervision from wikiHow}
\label{sec:distant}

The wikiHow repository contains high-quality articles describing the sequence of individual steps needed for the completion of a wide variety of practical tasks. Formally, we refer to wikiHow as a knowledge base $\mathbb{B}$ containing textual step descriptions for $T$ tasks:
$\mathbb{B} = \{y^{(1)}_1,...,y^{(1)}_{S_1}, \hdots, y^{(T)}_1,...,y^{(T)}_{S_T}\}$, where $y^{(t)}_s$ represents the language-based description of step $s$ for task $t$, and $S_t$ is the number of steps involved for the execution of task $t$. We view an instructional video $x$ as a sequence of $L$ segments $\{x_1,...,x_l,..,x_L\}$, with each segment $x_l$ consisting of $F$ RGB frames having spatial resolution $H \times W$, i.e., $x_l \in \mathbb{R}^{H \times W \times 3 \times F}$. Each video is accompanied by a paired sequence of text sentences $\{a_1,...,a_l,..,a_L\}$ obtained by applying ASR to the audio narration. We note that the narration $a_l$ can be quite noisy due to ASR errors. Furthermore, it may describe the step being executed only implicitly, e.g., by referring to secondary aspects. An example is given in Fig.~\ref{illustration}, where the ASR in the second segment describes the type of screws rather than the action of tightening the screws, while the last segment refers to the the tone confirmation of the air conditioner being activated rather than the plugging of the cord into the outlet.
The idea of our approach is to leverage the knowledge base $\mathbb{B}$ to de-noise the narration $a_l$ and to convert it into a supervisory signal that is more directly related to the steps represented in segments of the video. We achieve this goal through the framework of distant supervision, which we apply to approximate the unknown conditional distribution $P(y^{(t)}_s|x_l)$ over the steps executed in the video,  {\em without} any form of manual labeling. To approximate this distribution we employ a textual similarity measure $\mathcal{S}$ between $y^{(t)}_s$ and $a_l$ :

\begin{equation}
    P(y^{(t)}_s|x_l) \approx \frac{\exp{(\mathcal{S} (a_l,y^{(t)}_s))}}{\sum_{t,s} \exp{(\mathcal{S} (a_l,y^{(t)}_s))}}.
\end{equation}

The textual similarity $\mathcal{S}$ is computed as a dot product between language embeddings 
\begin{equation}
\mathcal{S} (a_l,y^{(t)}_s) =  e(a_l)^\top \cdot e(y^{(t)}_s)
\end{equation}
where $e(a_l),  e(y^{(t)}_s) \in \mathbb{R}^d$ and $d$ is the dimension of the language embedding space. The underlying intuition of our approach is that, compared to the noisy and unstructured narration $a_l$, the distribution $P(y^{(t)}_s|x_l)$ provides a more salient supervisory signal for training models to recognize individual steps of procedural activities in video. The last row of Fig.~\ref{illustration} shows the steps in the knowledge base having highest conditional probability given the ASR text. We can see that, compared to the ASR narrations, the step sentences provide a more fitting description of the step executed in each segment. Our key insight is that we can leverage modern language models to reassign noisy and imprecise speech transcriptions into the clean and informative step descriptions of our knowledge base. Beyond this qualitative illustration (plus additional ones available in the supplementary material), our experiments provide quantitative evidence of the benefits of training video models by using $P(y^{(t)}_s|x_l)$ as supervision as opposed to the raw narration. 


\subsection{Learning Step Embeddings from Unlabeled Video}
\label{sec:losses}

We use the approximated distribution $P(y^{(t)}_s|x_l)$ as the supervision to learn a video representation $f(x_l) \in \mathbb{R}^{d}$.
We consider three different training objectives for learning the video repesentation $f$: (1) step classification, (2) distribution matching, and (3) step regression. 

\noindent \textbf{Step Classification.} Under this learning objective, we first train a step classification model $\mathcal{F}_C: \mathbb{R}^{H \times W \times 3 \times F} \longrightarrow [0,1]^S$ to classify each video segment into one of the $S$ possible steps 
in the knowledge base $\mathbb{B}$, where $S = \sum_t S_t$.  
Specifically, let $t^*,s^*$ be the indices of the step in $\mathbb{B}$ that best describes segment $x_l$ according to our target distribution, i.e., 
\begin{equation}
t^*,s^* = \arg \max _{t,s} P(y^{(t)}_s|x_l). 
\end{equation}
Then, we use the standard cross-entropy loss to train $\mathcal{F}_C$ to classify video segment $x_l$ into class $(t^*,s^*)$:
\begin{equation}
    \min _\theta -\log\left( \left[ \mathcal{F}_C (x_l; \theta)  \right]_{(t^*,s^*)} \right)
\end{equation}
where $\theta$ denotes the learning parameters of the video model. The model uses a softmax activation function in the last layer to define a proper distribution over the steps, such that $\sum_{t,s} \left[ \mathcal{F}_C (x_l; \theta)  \right]_{(t,s)}  = 1$. Although here we show the loss for one segment $x_l$ only, in practice we optimize the objective by averaging over a mini-batch of video segments sampled from the entire collection in each iteration. After learning, we use $\mathcal{F}_C (x_l)$ as a feature extractor to capture step-level information from new video segments. Specifically, we use the second-to-last layer of $\mathcal{F}_C (x_l)$ (before the softmax function) as the step embedding representation $f(x_l)$ for classification of procedural activities in long videos. 

\noindent \textbf{Distribution Matching.} Under the objective of Distribution Matching, we train the step classification model $\mathcal{F}_{C}$ to minimize the KL-Divergence between the predicted distribution $\mathcal{F}_{C}(x_l)$ and the target distribution $P(y^{(t)}_s|x_l)$:
\begin{equation} 
    \min _\theta \sum_{t,s} P(y^{(t)}_s|x_l) \log \frac{P(y^{(t)}_s|x_l)}{\left[ \mathcal{F}_{C} (x_l; \theta)  \right]_{(t,s)}}.
\end{equation}
Due to the large step space ($S=10,588$), in order to effectively optimize this objective we empirically found it beneficial to use only the top-$K$ steps in $P(y^{(t)}_s|x_l)$, with the probabilities of the other steps set to zero.

\noindent \textbf{Step Regression.} Under Step Regression, we train the video model to predict the language embedding $e(y^{(t^*)}_{s^*}) \in \mathbb{R}^d$ associated to the pseudo ground-truth step $(t^*,s^*)$. Thus, in this case the model is a regression function to the language embedding space, i.e., $\mathcal{F}_R: \mathbb{R}^{H \times W \times 3 \times F} \longrightarrow \mathbb{R}^d$. We follow~\cite{miech2020end} and use the NCE loss as the objective:
\begin{equation} 
    \min _\theta - \log \frac{\exp \left( e(y^{(t^*)}_{s^*})^\top \mathcal{F}_R (x_l; \theta) \right)}{\sum_{(t,s) \neq (t^*,s^*)} \exp \left( e(y^{(t)}_s)^\top \mathcal{F}_R (x_l; \theta) \right) }
\end{equation}
Because $\mathcal{F}_R(x_l)$ is trained to predict the language representation of the step, we can directly use its output as step embedding representation for new video segments, i.e., $f(x_l) = \mathcal{F}_R(x_l)$.

\subsection{Classification of Procedural Activities}
In this subsection we discuss how we can leverage our learned step representation to recognize fine-grained procedural activities in long videos spanning up to several minutes. Let $x'$ be a new input video consisting of a sequence of $L'$ segments $x'_l \in \mathbb{R}^{H \times W \times 3 \times F}$ for $l=1, \hdots, L'$.  The intuition is that we can leverage our pretrained step representation to describe the video as a sequence of step embeddings. Because our step embeddings are trained to reveal semantic information about the individual steps executed in the segments, we use a transformer~\cite{vaswani2017attention} $\mathcal{T}$  to model dependencies over the steps and to classify the procedural activity: $\mathcal{T}(f(x'_1), \hdots, f(x'_{L'}))$. Since our objective is to demonstrate the effectiveness of our step representation $f$, we choose $\mathcal{T}$  to include a single transformer layer, which is sufficient to model sequential dependencies among the steps and avoids making the classification model overly complex. We refer to this model as the ``Basic Transformer.''

We also demonstrate that our step embeddings enable further beneficial information transfer from the knowledge base $\mathbb{B}$ to improve the classification of procedural activities during inference. The idea is to adopt a retrieval approach to find for each segment $x'_l$ the step $y_{s'}^{t'} \in \mathbb{B}$ that best explains the segment according to the pretrained video model $\mathcal{F} (x'_l; \theta)$. For the case of Step Classification and Distribution Matching, where we learn a classification model $\mathcal{F}_C (x'_l; \theta) \in [0,1]^S$, we simply select the step class yielding the maximum classification score:
\begin{equation}
t',s' = \arg \max _{t,s} [\mathcal{F}_C (x'_l; \theta)]_{(t,s)}.
\end{equation}
In the case of Step Regression, since $\mathcal{F}_R (x'_l; \theta)$ generates an output in the language space, we can choose the step that has maximum language embedding similarity:
\begin{equation}
\label{mmt}
t',s' = \arg \max _{t,s} e(y_s^{(t)})^\top \mathcal{F}_R (x'_l; \theta).
\end{equation}
Let $\hat{y}(x'_l)$ denote the step description assigned through this procedure, i.e., $\hat{y}(x'_l) = y_{s'}^{t'}$.

Then, we can incorporate the knowledge retrieved from $\mathbb{B}$ for each segment in the input provided to the transformer, together with the step embeddings extracted from the video:
\begin{equation}
    \mathcal{T}(f(x'_1), e(\hat{y}(x'_1)), f(x'_2), e(\hat{y}(x'_2)), ..., f(x'_{L'}), e(\hat{y}(x_{L'}))).
\end{equation}
This formulation effectively trains the transformer to fuse a representation consisting of video features and step embeddings from the knowledge base to predict the class of the procedural activity. We refer to this variant as  ``Transformer w/ KB Transfer''. 

\subsection{Step Forecasting}
We note that we can easily modify our proposed classification model to address forecasting tasks that require long-term analysis over a sequence of steps to predict future activity. One such problem is the task of ``next-step anticipation'' which we consider in our experiments. Given as input a video spanning $M$ segments, $\{x_1, \hdots, x_M\}$, the objective is to predict the step executed in the {\em unobserved} $(M+1)$-th segment. To address this task we train the transformer on the sequence step embeddings extracted from the $M$ observed segments. In the case of Transformer w/ KB Transfer, for each input segment $x'_l$, we include $f(x'_l)$ 
but also $e(y_{s'+1}^{t'})$, i.e., the embedding of the step immediately after the step matched in the knowledge base. This effectively provides the transformer with information about the likely future steps according to the knowledge base.

\subsection{Model Design}

We use MPNet~\cite{mpnet} as the language model to extract 768-dimensional language embeddings for both the ASR sentences and the step descriptions in wikiHow articles. MPNet (paraphrase-mpnet-base-v2) is at the time of writing (August, 2021) ranked first by Sentence Transformers~\cite{sbert}, based on performance across 14 language retrieval tasks~\cite{reimers-2019-sentence-bert}. The similarity between two embedding vectors is chosen to be the dot product between the two vectors. 

We choose as video model the TimeSformer architecture~\cite{bertasius2021space}. Starting from a configuration of ViT initialized with ImageNet-21K ViT pretraining~\cite{dosovitskiy2020image}, we train TimeSformer on HowTo100M using clips of 8 frames uniformly sampled from time-spans of 8 seconds. The evaluations in our experiments are carried out by learning the step representation on HowTo100M (without manual labels) and by assessing the performance of our embeddings on smaller-scale downstream datasets where task and/or step manual annotations are available. To perform classification of multi-step activities on these downstream datasets we use a single transformer layer~\cite{vaswani2017attention} trained on top of our fixed embeddings. We use this shallow long-term model without finetuning in order to directly measure the value of the representation learned via distant supervision from the unlabeled instructional videos. We refer the reader to the supplementary material for additional implementation details.

\begin{figure}[t]
\begin{center}
\includegraphics[width=0.8\linewidth]{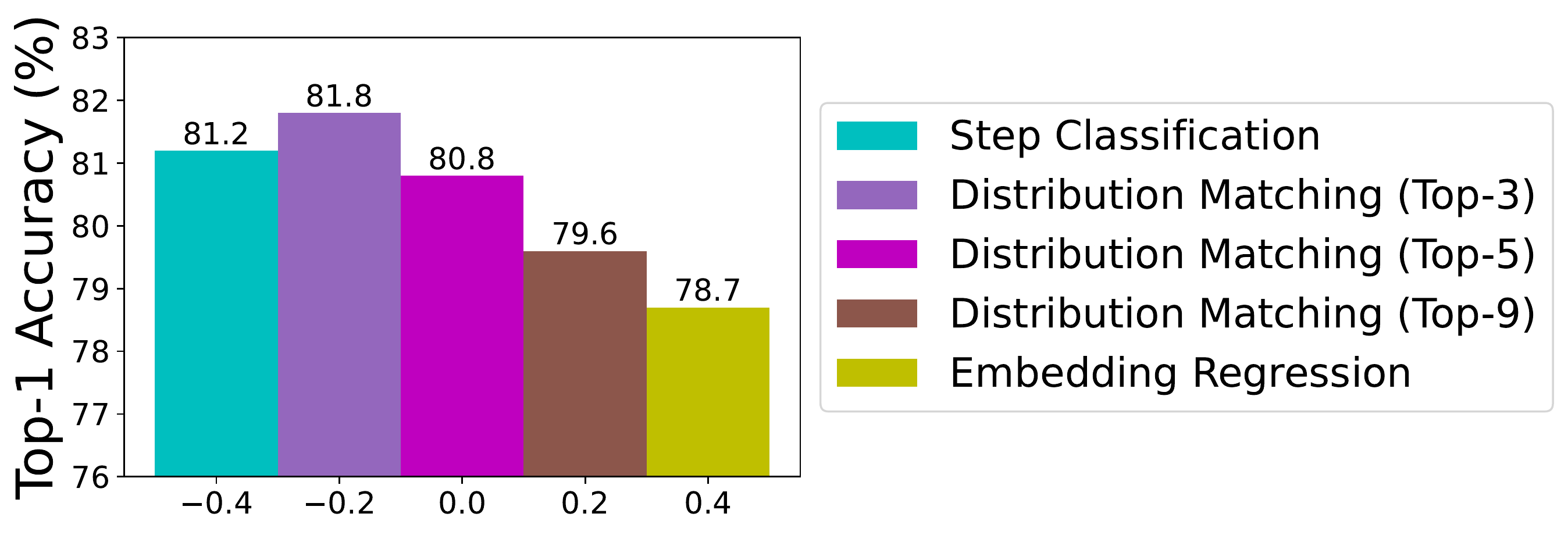}\vspace{-.5cm}
\end{center}
\caption{Accuracy of classifying procedural activities in COIN using three different distant supervision objectives.\vspace{-.2cm}
}
\label{fig:ablateloss}
\end{figure}



\section{Experiments}

\subsection{Datasets and Evaluation Metrics}
\noindent \textbf{Pretraining.} HowTo100M (HT100M)~\cite{miech2019howto100m} includes over 1M long instructional videos split into about 120M video clips in total. We use the complete HowTo100M dataset only in the final comparison with the state-of-the-art (sec.~\ref{sec:sota}). In the ablations, in order to reduce the computational cost, we use a smaller subset corresponding to the collection of 80K long videos defined by Bertasius \etal~\cite{bertasius2021space}.

\noindent \textbf{Classification of Procedural Activities.} Performance on this task is evaluated using two labeled datasets: COIN~\cite{tang2019coin,tang2020comprehensive} and Breakfast~\cite{breakfast}. COIN contains about 11K instructional videos representing 180 tasks (i.e., classes of procedural activities). Breakfast~\cite{breakfast} contains 1,712 videos for 10 complex cooking tasks. 
In both datasets, each video is manually annotated with a label denoting the task class. We use the standard splits~\cite{tang2020comprehensive,hussein2019timeception} for these two datasets and measure performance as task classification accuracy.

\noindent \textbf{Step Classification.} It requires classifying the step observed in a single video segment (without history), which is a good testbed to evaluate the effectiveness of our step embeddings. To evaluate methods on this problem, we use the step annotations from COIN, corresponding to a total of 778 step classes representing parts of tasks. The steps are manually annotated within each video with temporal boundaries and step class labels. 
Classification accuracy~\cite{tang2020comprehensive} of a linear classifier (Linear Acc) is used as the metric. 

\noindent \textbf{Step Forecasting.} We also use step annotations available in COIN. The objective is to predict the class of the step in the next segment given as input the sequence of observed video segments up to that step (excluded). 
Note that there is a substantial temporal gap (21 seconds on average) between the end of the last observed segment and the start of the step to be predicted. This makes the problem quite challenging and representative of real-world conditions. We set the history to contain at least one step. We use classification accuracy of the predicted step as the evaluation metric.

\begin{figure}[t]
\begin{center}
\includegraphics[width=0.9\linewidth]{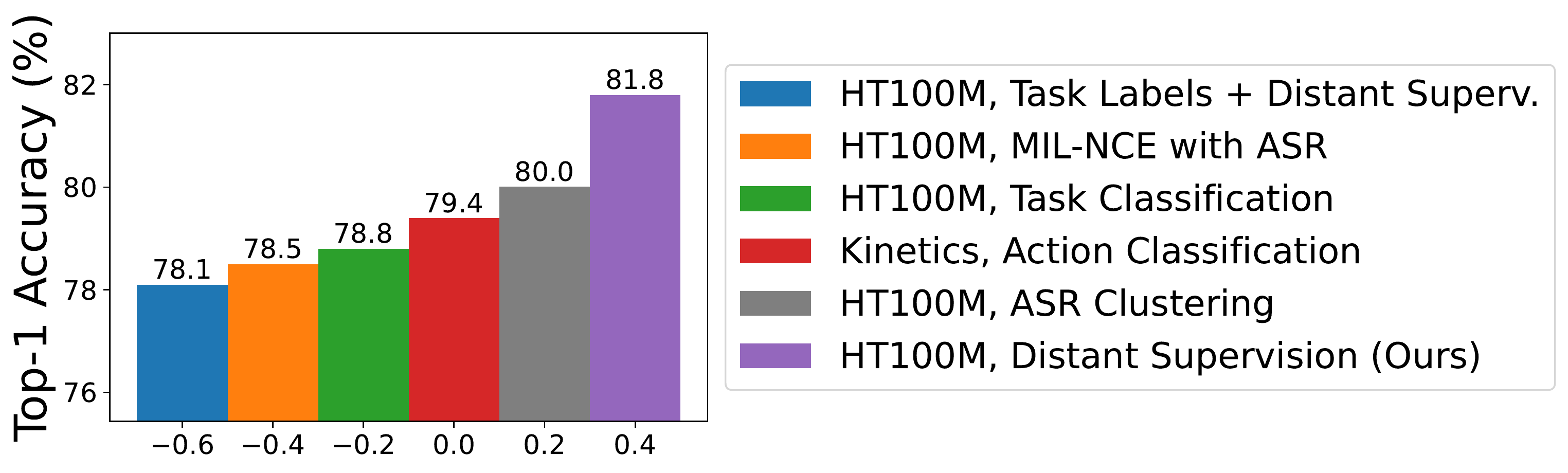}\vspace{-.5cm}
\end{center}
\caption{Accuracy of procedural activity classification on COIN using video representations learned with different  supervisions.\vspace{-.2cm}}

\label{fig:step}
\end{figure}

\noindent \textbf{Egocentric Activity Recognition.} EPIC-KITCHENS-100~\cite{damen2020epic} is a large-scale egocentric video dataset. It consists of 100 hours of first-person videos, showing humans performing a wide range of procedural activities in the kitchen. The dataset includes manual annotations of 97 verbs and 300 nouns in manually-labeled video segments. We follow the standard protocol~\cite{damen2020epic} to train and evaluate our models.

 \begin{table*}[th]
\centering
\footnotesize
\begin{tabular}{lllll}
\hline
Segment Model   & Pretraining Supervision & Pretraining Dataset  & Linear Acc (\%)
\\ \hline 
TSN (RGB+Flow)~\cite{tang2020comprehensive}  & Supervised: action labels & Kinetics      & 36.5* \\ 

S3D~\cite{miech2020end}  & Unsupervised: MIL-NCE on ASR  & HT100M      & 37.5* \\ 
\hline \hline
ClipBERT~\cite{lei2021less} & Supervised: captions & COCO + Visual Genome &  30.8\\
VideoCLIP~\cite{xu2021videoclip} & Unsupervised: NCE on ASR  & HT100M      & 39.4 \\ 

 SlowFast~\cite{feichtenhofer2019slowfast}  & Supervised: action labels  & Kinetics   & 32.9 \\
 TimeSformer~\cite{bertasius2021space}   & Supervised: action labels & Kinetics      & 48.3 \\
  TimeSformer~\cite{bertasius2021space}   & Unsupervised: $k$-means on ASR & HT100M      & 46.5 \\
 \textbf{TimeSformer}   & \textbf{Unsupervised: distant supervision (ours)} & HT100M      & \textbf{54.1} \\
\hline
\end{tabular}
\caption{Comparison to the state-of-the-art for \textbf{step classification} on the COIN dataset. * indicates results by finetuning on COIN.
}

\label{sota:step}
\end{table*}


\begin{table*}[t]
\footnotesize
\begin{tabular}{lllll}
\hline
Long-term Model & Segment Model  & Pretraining Supervision & Pretraining Dataset  & Acc (\%)
\\ \hline 
TSN (RGB+Flow)~\cite{tang2020comprehensive} & Inception~\cite{szegedy2016rethinking} & Supervised: action labels & Kinetics      & 73.4* \\ 

Basic Transformer & S3D~\cite{miech2020end}  & Unsupervised: MIL-NCE on ASR  & HT100M     & 70.2* \\ 
\hline \hline
Basic Transformer & ClipBERT~\cite{lei2021less} & Supervised: captions & COCO + Visual Genome & 65.4 \\
Basic Transformer & VideoCLIP~\cite{xu2021videoclip} & Unsupervised: NCE on ASR  & HT100M      &  72.5\\ 

Basic Transformer & SlowFast~\cite{feichtenhofer2019slowfast}  & Supervised: action labels  & Kinetics      & 71.6 \\
Basic Transformer & TimeSformer~\cite{bertasius2021space}   & Supervised: action labels & Kinetics      & 83.5 \\
Basic Transformer & TimeSformer~\cite{bertasius2021space}   & Unsupervised: $k$-means on ASR & HT100M      & 85.3 \\
\textbf{Basic Transformer} & \textbf{TimeSformer}   & \textbf{Unsupervised: distant supervision (ours)} & HT100M     &  \textbf{88.9} \\
\textbf{Transformer w/ KB Transfer} & \textbf{TimeSformer}   & \textbf{Unsupervised: distant supervision (ours)} & HT100M     & \textbf{90.0}
\\
\hline
\end{tabular}
\centering
\caption{Comparison to the state-of-the-art for the \textbf{classification of procedural activities} on the COIN dataset.\vspace{-.3cm}}
\label{sota:task_coin}
\end{table*}

\subsection{Ablation Studies}

We begin by studying how different design choices in our framework affect the accuracy of task classification on COIN using the basic Transformer as our long-term model. 

\subsubsection{Different Training Objectives}

Fig.~\ref{fig:ablateloss} shows the accuracy of COIN task classification using the three distant supervision objectives presented in Sec.~\ref{sec:losses}. Distribution Matching and Step Classification achieve similar performance, while Embedding Regression produces substantially lower accuracy. Based on these results we choose Distribution Matching (Top-3) as our learning objective for all subsequent experiments. 


\subsubsection{Comparing Different Forms of Supervision}

In Fig.~\ref{fig:step}, we compare the results of different pretrained video representations for the problem of classifying procedural activities on the COIN dataset. We include as baselines several representations learned on the same subset of HowTo100M as our step embeddings, using the same TimeSformer as video model. MIL-NCE~\cite{miech2020end} performs contrastive learning between the video and the narration obtained from ASR. The baseline (HT100M, Task Classification) is a representation learned by training TimeSformer as a classifier using as classes the task ids available in HowTo100M. The task ids are defined by the keywords used to find the video on YouTube. The baseline (HT100M, Task Labels + Distant Superv.) uses the task ids to narrow down the potential steps considered by distant supervision (only wikiHow steps corresponding to the task id of the video are considered). We also include a representation obtained by training TimeSformer on the fully-supervised Kinetics-400 dataset~\cite{Kinetics}. Finally, to show the benefits of distant supervision, we run $k$-means clustering on the language embeddings of ASR sentences using the same number of clusters as the steps in wikiHow (i.e., $k=S=10,588$), and then train the video model using the cluster ids as supervision. 

We observe several important results in Fig.~\ref{fig:step}. First, our distant supervision achieves an accuracy gain of $3.3\%$ over MIL-NCE with ASR. 
This suggests that our distant supervision framework provides more explicit supervision to learn step-level representations compared to using directly the ASR text. This is further confirmed by the performance of ASR Clustering, which is $1.7\%$ lower than that obtained by leveraging the wikiHow knowledge base.

Moreover, our step-level representation outperforms by $3\%$ the weakly-supervised task embeddings (Task Classification) and does even better (by $2.4\%$) than the video representation learned with full supervision from the large-scale Kinetics dataset. This is due to the fact that steps typically involve multiple atomic actions. For example, about $85\%$ of the steps consist of at least two verbs. Thus, our step embeddings capture a higher-level representation than those based on traditional atomic action labels. 

Finally, using the task ids to restrict the space of step labels considered by distant supervision produces the worst results. This indicates that the task ids are quite noisy and that our approach leveraging relevant steps from other tasks can provide more informative supervision. These results further confirm the superior performance of distantly supervised step annotations over existing task or action labels to train representations for classifying procedural activities.

 \begin{table*}[t]
  
\begin{center}
\footnotesize
\begin{tabular}{llllll}
\hline
Long-term Model & Segment Model  & Pretraining Supervision & Pretraining Dataset  & Acc (\%)
\\ \hline 
Timeception~\cite{hussein2019timeception} & 3D-ResNet~\cite{wang2017non} & Supervised: action labels & Kinetics      & 71.3 \\  
VideoGraph~\cite{hussein2019videograph} & I3D~\cite{Kinetics}  & Supervised: action labels & Kinetics     &  69.5 \\  
GHRM~\cite{zhou2021graph} & I3D~\cite{Kinetics}  & Supervised: action labels & Kinetics     &  75.5 \\  
\hline \hline
Basic Transformer & S3D~\cite{miech2020end}  & Unsupervised: MIL-NCE on ASR  & HT100M      & 74.4 \\ 
Basic Transformer & SlowFast~\cite{feichtenhofer2019slowfast}  & Supervised: action labels  & Kinetics      & 76.1 \\
Basic Transformer & TimeSformer~\cite{bertasius2021space}   & Supervised: action labels & Kinetics      & 81.1 \\
Basic Transformer & TimeSformer~\cite{bertasius2021space}   & Unsupervised: $k$-means on ASR & HT100M      & 81.4 \\
\textbf{Basic Transformer} & \textbf{TimeSformer}   & \textbf{Unsupervised: distant supervision (ours)} & HT100M     &  \textbf{88.7} \\
\textbf{Transformer w/ KB Transfer} & \textbf{TimeSformer}   & \textbf{Unsupervised: distant supervision (ours)} & HT100M     & \textbf{89.9}\\
\hline 
\end{tabular}
\end{center}
\vspace{-.4cm}
\caption{Comparison to the state-of-the-art for the problem of classifying procedural activities on the Breakfast dataset. 
}
\vspace{-.2cm}
\label{sota:task_bf}
\end{table*}

 \begin{table*}[t]
\centering
\footnotesize
\begin{tabular}{lllllll}
\hline
Long-term Model & Segment Model  & Pretraining Supervision & Pretraining Dataset  & Acc (\%)
\\ \hline 

Basic Transformer & S3D~\cite{miech2020end}  & Unsupervised: MIL-NCE on ASR  & HT100M     & 28.1 \\ 
Basic Transformer & SlowFast~\cite{feichtenhofer2019slowfast}  & Supervised: action labels  & Kinetics     & 25.6 \\
Basic Transformer & TimeSformer~\cite{bertasius2021space}   & Supervised: action labels & Kinetics      & 34.7 \\
Basic Transformer & TimeSformer~\cite{bertasius2021space}   & Unsupervised: $k$-means on ASR & HT100M      & 34.0 \\
\textbf{Basic Transformer} & \textbf{TimeSformer}   & \textbf{Unsupervised: distant supervision (ours)} & HT100M     & \textbf{38.2} \\
\textbf{Transformer w/ KB Transfer} & \textbf{TimeSformer}   & \textbf{Unsupervised: distant supervision (ours)} & HT100M     &  \textbf{39.4} \\
\hline
\end{tabular}
\vspace{-.1cm}
\caption{Accuracy of different methods on the step forecasting task using the COIN dataset.}
\vspace{-.2cm}
\label{sota:anticipate}
\end{table*}

\begin{table*}[!h]
  
\begin{center}
\footnotesize
\begin{tabular}{llllllll}
\hline
Segment Model   & Pretraining Supervision & Pretraining Dataset &  Action (\%) & Verb (\%) & Noun (\%)
\\ \hline 
TSN~\cite{TSN}  & -  & -     &  33.2 & 60.2 & 46.0  \\
TRN~\cite{zhou2018temporal}  & -  & -      & 35.3 & 65.9 & 45.4  \\
TBN~\cite{kazakos2019epic}  & -  & -      & 36.7 & 66.0 & 47.2 \\
MoViNet~\cite{movinetsarxiv} & - & - & \textbf{47.7} & \textbf{72.2} & 57.3 \\
TSM~\cite{lin2018temporal}  & Supervised: action labels  & Kinetics     & 38.3 & 67.9 & 49.0  \\
SlowFast~\cite{feichtenhofer2019slowfast}  & Supervised: action labels  & Kinetics      & 38.5 & 65.6 & 50.0  \\
\hline
ViViT-L~\cite{arnab2021vivit}  & Supervised: action labels  & Kinetics      & 44.0 & 66.4 & 56.8  \\
\hline \hline
TimeSformer~\cite{bertasius2021space}   & Supervised: action labels & Kinetics      & 42.3 & 66.6 & 54.4  \\
\textbf{TimeSformer}   & \textbf{Unsupervised: distant supervision (ours)} & HT100M      & {44.4} & {67.1} & \textbf{58.1} \\
\hline
\end{tabular}
\end{center}
\vspace{-.3cm}
\caption{Comparison to the state-of-the-art for classification of first-person videos using the EPIC-KITCHENS-100 dataset.\vspace{-.3cm}}
\label{sota:epic}
\end{table*}

\subsection{Comparisons to the State-of-the-Art}
\label{sec:sota}

\subsubsection{Step Classification}

We study the problem of step classification as it directly measures whether the proposed distant supervision framework provides a useful training signal for recognizing steps in video. For this purpose, we use our distantly supervised model as a frozen feature extractor to extract step-level embeddings for each video segment and then train a linear classifier to recognize the step class in the input segment. 

Table~\ref{sota:step} shows that our distantly supervised representation achieves the best performance and yields a large gain over several strong baselines. Even on this task, our distant supervision produces better results compared to a video representation trained with fully-supervised action labels on Kinetics. The significant gain ($7.6\%$) over ASR clustering again demonstrates the importance of using wikiHow knowledge. Finally, our model achieves strong gains over previously reported results on this benchmark based on different backbones, including results obtained by finetuning and using optical flow as an additional modality~\cite{tang2020comprehensive}. 



\subsubsection{Classification of Procedural Activities}
Table~\ref{sota:task_coin} and Table~\ref{sota:task_bf} show accuracy of classifying procedural activities in long videos on the COIN and Breakfast dataset, respectively. Our model outperforms all previous works on these two benchmarks. For this problem, the accuracy gain on COIN over the representations learned with Kinetics action labels has become even larger ($6.5\%$) compared to the improvement achieved for step classification ($5.8\%$). This indicates that the distantly supervised representation is indeed highly suitable for recognizing long procedural activities.
We also observe a substantial gain ($8.8\%$) over the Kinetics baseline for the problem of recognizing complex cooking activities in the Breakfast dataset. As GHRM provided also the result obtained by finetuning the feature extractor on the Breakfast benchmark (89.0\%), we measured the accuracy achieved by finetuning our model and observed a large gain: 91.6\%. We also tried replacing the basic transformer with Timeception as the long-term model. Timeception trained on features learned with action labels from Kinetics gives an accuracy of $79.4\%$. This same model trained on our step embeddings achieves an accuracy of $83.9\%$. The large gain confirms the superiority of our representation for this task and suggests that our features can be effectively plugged in different long-term models. 

\subsubsection{Step Forecasting}
Table~\ref{sota:anticipate} shows that our learned representation and a shallow transformer can be used to forecast the next step very effectively. 
Our representation outperforms the features learned with Kinetics action labels by $3.5\%$. When the step order knowledge is leveraged by stacking the embeddings of the possible next steps, the gain is further improved to $4.7\%$. This shows once more the benefits of incorporating information from the wikiHow knowledge base.

\subsubsection{Egocentric Video Understanding}

Recognition of activities in EPIC-KITCHENS-100~\cite{damen2020epic} is a relevant testbed for our model since first-person videos in this dataset capture diverse procedural activities from daily human life. To demonstrate the generality of our distantly supervised approach, we finetune our pretrained model for the task of noun, verb, and action recognition in egocentric videos. For comparison purposes, we also include the results of finetuning the same model pretrained on Kinetics-400 with manually annotated action labels. Table~\ref{sota:epic} shows that the finetuning of our distantly supervised model  outperforms all prior works, with the only exception of MoViNet~\cite{movinetsarxiv} which achieves higher accuracies for Action and Verb but not for Noun. This provides further evidence about the transferability of our models to other tasks.

\section{Conclusion}
In this paper, we introduce a distant supervision framework that leverages a textual knowledge base (wikiHow) to effectively learn step-level video representations from instructional videos. We demonstrate the value of the representation on step classification, long procedural video classification, and step forecasting. We further show that our distantly supervised model generalizes well to egocentric video understanding. 

\section*{Acknowledgments}
Thanks to Karl Ridgeway, Michael Iuzzolino, Jue Wang, Noureldien Hussein, and Effrosyni Mavroudi for valuable discussions.  

{\small
\bibliographystyle{ieee_fullname}
\bibliography{egbib}
}

\clearpage
\appendix

\section{Implementation Details}

Our implementation uses the wikiHow articles collected and processed by Koupaee and Wang~\cite{koupaee2018wikihow}, where each article has been parsed into a title and a list of step descriptions. We use a total of $S=10,588$ steps collected from the $T=1059$ tasks used in the evaluation of Bertasius \etal~\cite{bertasius2021space}. This represents the subset of wikiHow tasks that have at least 100 video samples in the HowTo100M dataset. We note that the HowTo100M videos were collected from YouTube~\cite{youtube} by using the wikiHow titles as keywords for the searches. Thus, each task of HowTo100M is represented in the knowledge base of wikiHow, except for tasks deleted or revised.

We implement our video model using the code base of TimeSformer~\cite{bertasius2021space}. All methods and baselines based on TimeSformer start from a configuration of ViT initialized with ImageNet-21K ViT pretraining~\cite{dosovitskiy2020image}. Each segment consists of 8 frames uniformly sampled from a time-span of 8 seconds. For pretraining, we sample segments according to the ASR temporal boundaries available in HowTo100M. If the time-span exceeds 8 seconds, we sample a segment randomly within it, otherwise we take the 8-second segment centered at the middle point. For our pretraining of TimeSformer on the whole set of HowTo100M videos, we use a configuration slightly different from that adopted in~\cite{bertasius2021space}. We use a batch size of 256 segments, distributed over 128 GPUs to accelerate the training process. The models are first trained with the same optimization hyper-parameter settings for 15 epochs as~\cite{bertasius2021space}. Then the models are trained with AdamW~\cite{loshchilov2018decoupled} for another 15 epochs, with an initial learning rate of 0.00005. The pretraining of our model took 55 hours using 128 GPUs. As a reference, Miech et al.~\cite{miech2020end} report that pretraining S3D with MIL-NCE on HowTo100M required 3 days with 64 8-core TPUs.

To perform classification of multi-step activities as well as step forecasting on downstream datasets we use a single basic transformer layer~\cite{vaswani2017attention} trained on top of our fixed embeddings. The transformer layer has 768 embedding dimensions and 12 heads. The step embeddings extracted with TimeSformer are augmented with learnable positional embeddings before being fed to the transformer layer. We train the transformer layer on sequences of 8 embedding vectors extracted from a series of 8 adjacent 8-second clips from the input video (spanning a total of 64 seconds).

For step classification, we train a simple linear classifier on embeddings extracted from individual segments of the downstream dataset. If the segment exceeds 8 seconds we sample the middle clip of 8 seconds, otherwise we use the given segment and sample 8 frames from it uniformly. 

For egocentric video classification on EPIC-KITCHENS-100, we follow the experimental setup described in~\cite{arnab2021vivit}, except that we sample 32 frames as input with a frame rate of 2 fps to cover a longer temporal span of 16 seconds.

For the downstream tasks of procedural activity recognition, step classification, and step anticipation, we train the extra layer(s) on top of the frozen step embedding representation for 75K iterations, starting with a learning rate of 0.005. The learning rate is scaled by 0.1 after 55K and 70K iterations, respectively. The optimizer is SGD. We ensemble predictions from 4, 3, and 4 temporal clips sampled from the input video for the three tasks, respectively. We follow ~\cite{tang2019coin,breakfast} to split the data sets into a training set and a test set on the COIN and the Breakfast dataset, respectively.

\section{Classification Results with Different Number of Transformer Layers}

In the main paper, we presented results for recognition of procedural activities using as classification model a single-layer Transformer trained on top of the video representation learned with our distant supervision framework. In Table~\ref{ablate:layers} we study the potential benefits of additional Transformer layers. We can see that additional Transformer layers in the classifier do not yield significant gains in accuracy. This suggests that our representation enables accurate classification of complex activities with a simple model and does not require additional nonlinear layers to achieve strong recognition performance. We also show the results without any transformer layers, 
by training a linear classifier on the average pooled or concatenated features from the pretrained TimeSformer. It has a substantially low results compared to using transformer layers for temporal modeling, which indicates that our step-level representation enables effective powerful temporal reasoning even with a simple model.

\section{Representation Learning with Different Video Backbones}

Although the experiments in our paper were presented for the case of TimeSformer as the video backbone, our distant supervision framework is general and can be applied to any video architecture. To demonstrate the generality of our framework, in this supplementary material we report results obtained with another recently proposed video model, ST-SWIN~\cite{wang2021long}, using ImageNet-1K pretraining as initialization.
We first train the model on HowTo100M using our distant supervision strategy and then evaluate the learned (frozen) representation on the tasks of step classification and procedural activity classification in the COIN dataset. Table~\ref{swin_step} and Table~\ref{swin} show the results for these two tasks. We also include results achieved with a video representation trained with full supervision on Kinetics as well as with video embeddings learned by $k$-means on ASR text. As we have already shown for the case of TimeSformer in the main paper, even for the case of the ST-SWIN video backbone, our distant supervision provides the best accuracy on both benchmarks, outperforming the Kinetics and the $k$-means baseline by substantial margins. This confirms that our distant supervision framework can work effectively with different video architectures.

\begin{table}[t]
\footnotesize
\begin{tabular}{p{1.6cm}p{2cm}p{2.7cm}}
\hline
\# Transformer Layers & Acc (\%) of Basic Transformer & Acc (\%) of Transformer w/ KB Transfer 
\\ \hline 
0 (Avg Pool) & 81.0 & n/a \\
0 (Concat) & 81.5 & n/a \\
1 & 88.9     & 90.0 \\
2  & 90.0     & 89.8 \\
3  & 89.3     &  90.4 \\
\hline
\end{tabular}
\centering
\caption{Effect of different number of Transformer layers in the classification model used to recognize procedural  activities in  the  COIN  dataset. The classifier is trained on top of the video representation learned with our distant supervision framework. 
}
\label{ablate:layers}
\end{table}

\begin{table*}[t]
\footnotesize
\begin{tabular}{llll}
\hline
Segment Model  & Pretraining Supervision & Pretraining Dataset  & Acc (\%)
\\ \hline 

 ST-SWIN   & Supervised: action labels & Kinetics      & 44.0 \\
 ST-SWIN   & Unsupervised: $k$-means on ASR & HT100M      & 44.8 \\
 \textbf{ST-SWIN}   & \textbf{Unsupervised: distant supervision (ours)} & HT100M     &  \textbf{50.3} \\
\hline
\end{tabular}
\centering
\caption{Comparison to baselines for the problem of step classification on the COIN dataset using ST-SWIN as video architecture.}
\label{swin_step}
\end{table*}

\begin{table*}[t]
\footnotesize
\begin{tabular}{lllll}
\hline
Long-term Model & Segment Model  & Pretraining Supervision & Pretraining Dataset  & Acc (\%)
\\ \hline 

Basic Transformer & ST-SWIN   & Supervised: action labels & Kinetics      & 79.6 \\
Basic Transformer & ST-SWIN   & Unsupervised: $k$-means on ASR & HT100M      & 82.4 \\
\textbf{Basic Transformer} & \textbf{ST-SWIN}   & \textbf{Unsupervised: distant supervision (ours)} & HT100M     &  \textbf{88.3} \\
\hline
\end{tabular}
\centering
\caption{Comparison to baselines for the problem of classifying procedural activities on the COIN dataset using ST-SWIN as video architecture.}
\label{swin}
\end{table*}

\section{Action Segmentation Results on COIN}
In the main paper, we use step classification on COIN as one of the downstream tasks to directly measure the quality of the learned step-level representations. We note that some prior works~\cite{ActBERT,miech2020end} used the step annotations in COIN to evaluate pretrained models for action segmentation. This task entails densely predicting action labels at each frame. Frame-level accuracy is used as the evaluation metric. 
We argue that step classification is a more relevant task for our purpose since we are interested in understanding the representational power of our features as step descriptors. Nevertheless, in order to compare to prior works, here we present results of using our step embeddings for action segmentation on COIN. 
Following previous work~\cite{ActBERT,miech2020end}, we sample adjacent non-overlapping 1-second segments from the long video as input to our model. We use our model pretrained on HowTo100M as a fixed feature extractor to obtain a representation for each of these segments. Then a linear classifier is trained to classify each segment into one of the 779 classes (778 steps plus the background class). Our method achieves a frame accuracy of $67.6\%$.  
The representation learned with full-supervision using action labels in Kinetics gives a substantially lower accuracy: $63.8\%$ with the same classification model as our method. The methods in~\cite{ActBERT,miech2020end} achieve an accuracy of $57.0\%$ and $61.0\%$, respectively. Using
the same linear setup as our model, VideoCLIP features~\cite{xu2021videoclip} pretrained on HowTo100M achieve an accuracy of $59.9\%$, i.e., $7.7\%$ lower than our representation.



\section{More Qualitative Results and Discussion}
\begin{figure*}[t]
\centering
\begin{subfigure}[t]{\linewidth}
    \centering
    \includegraphics[width=0.9\linewidth]{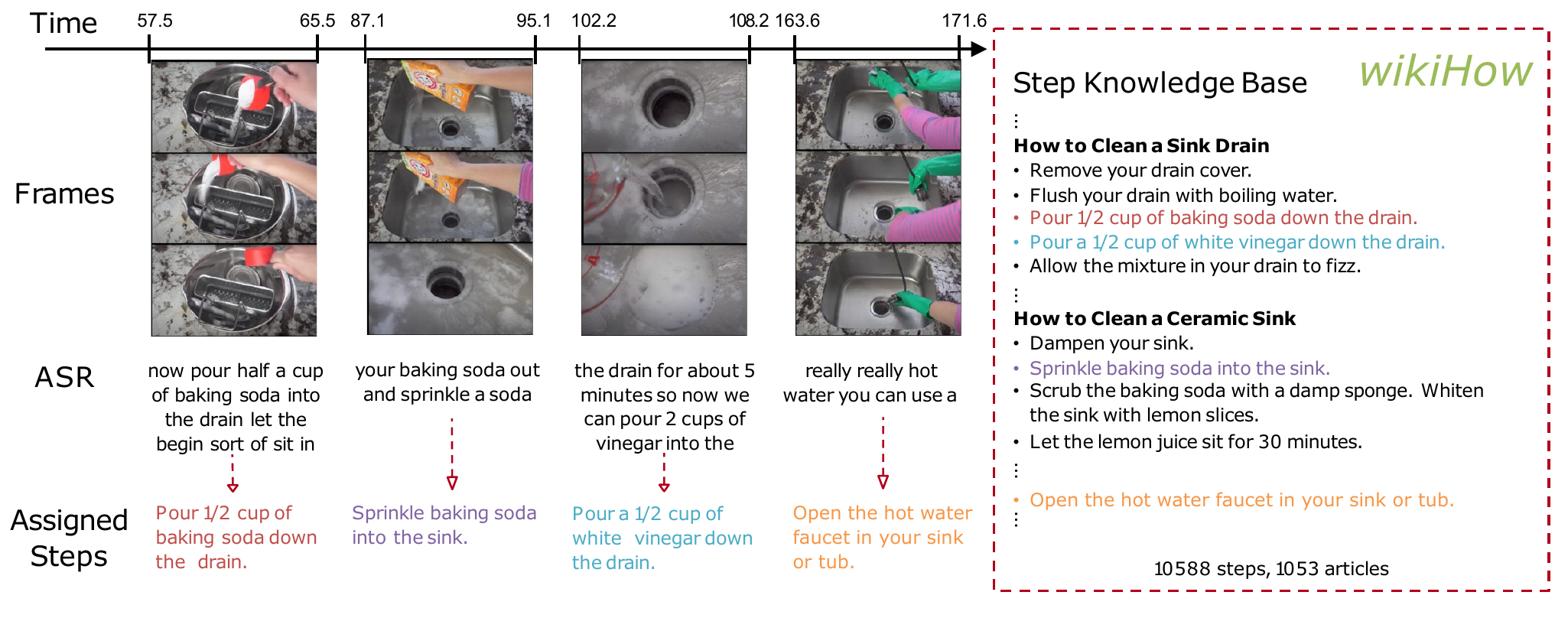}
    \caption{}
    \label{vis1:a}
\end{subfigure}
\begin{subfigure}[t]{\linewidth}
    \centering
    \includegraphics[width=0.9\linewidth]{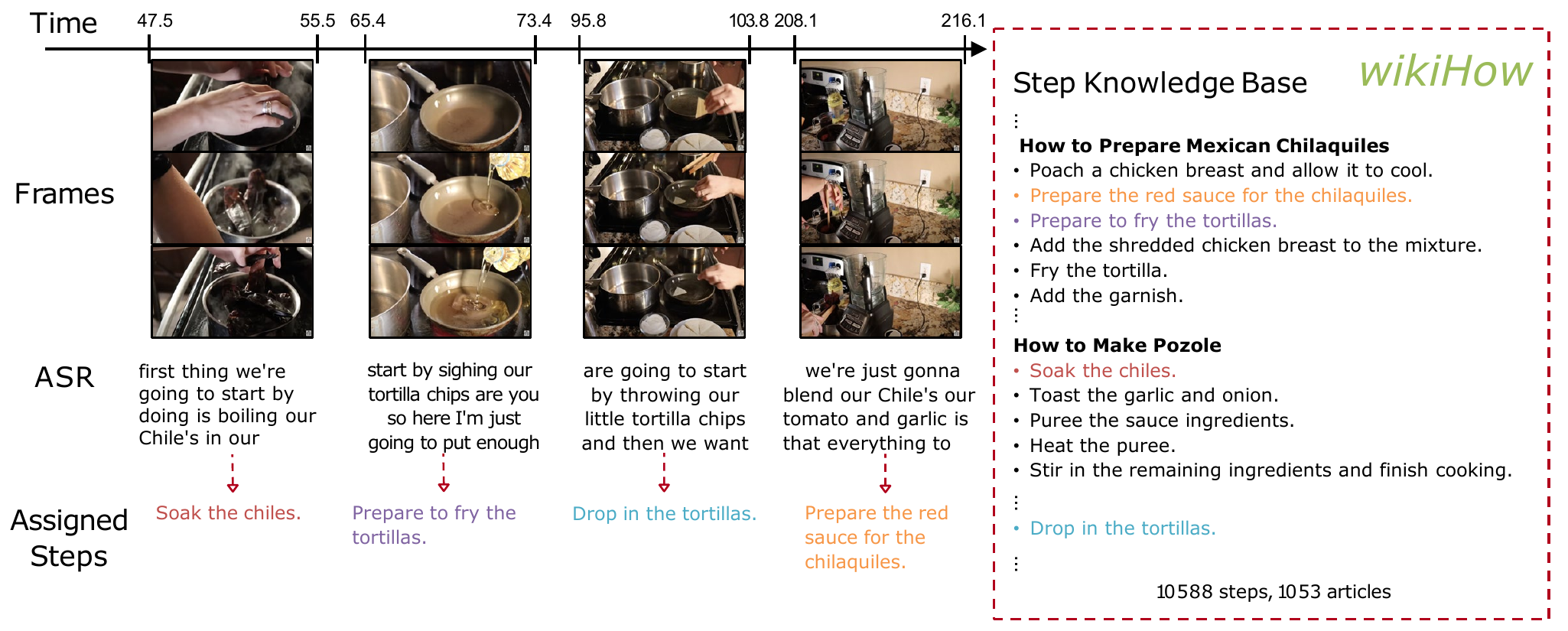}
    \caption{}
    \label{vis1:b}
\end{subfigure}
\begin{subfigure}[t]{\linewidth}
    \centering
    \includegraphics[width=0.9\linewidth]{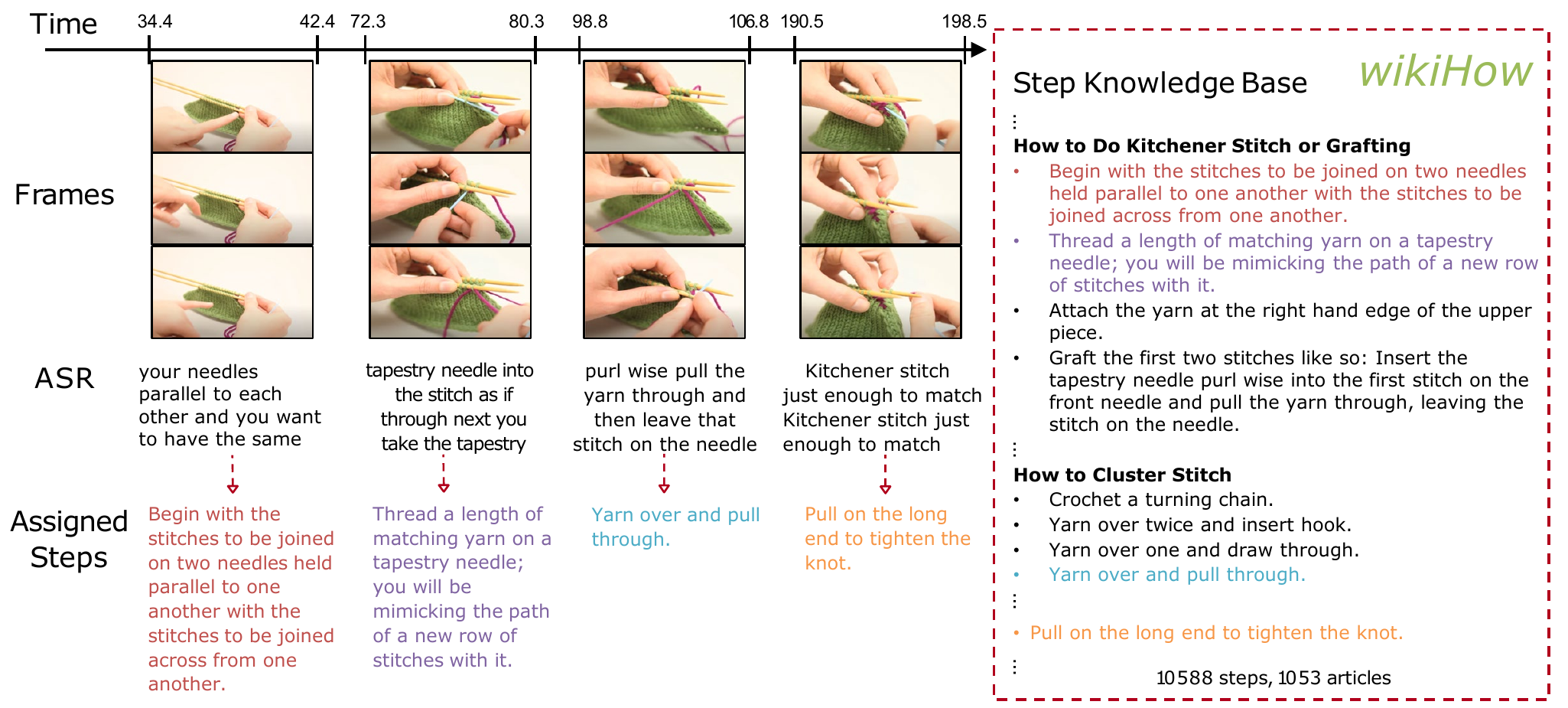}
    \caption{}
    \label{vis1:c}
\end{subfigure}
\caption{Visualization of steps assigned by our distant supervision method for three different video examples from the HowTo100M dataset. The assigned steps provide more expressive descriptions compared to the noisy ASR sentences.}
\label{vis1}
\end{figure*}

\begin{figure*}[t]
\centering
\begin{subfigure}[t]{\linewidth}
    \centering
    \includegraphics[width=0.8\linewidth]{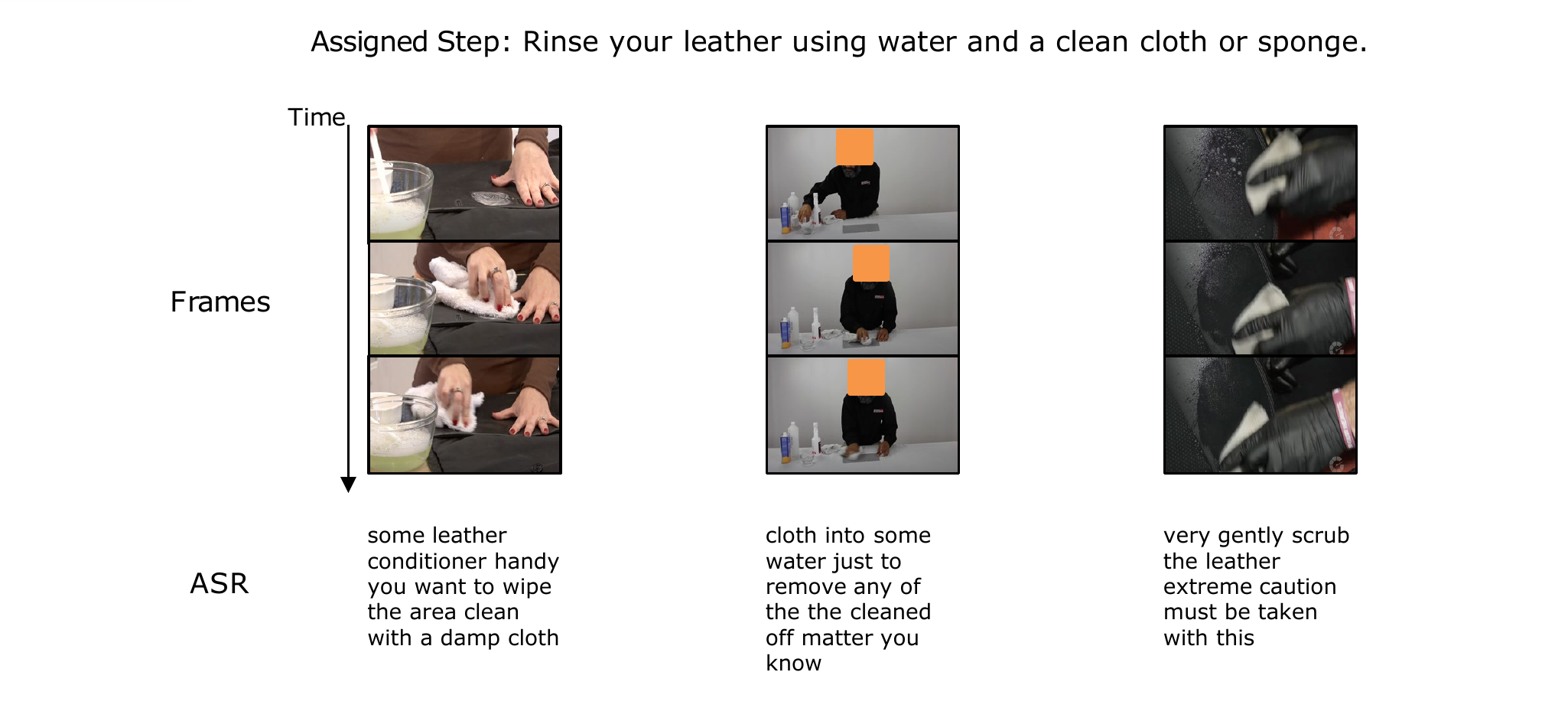}
    \caption{}
    \label{vis2:a}
\end{subfigure}
\begin{subfigure}[t]{\linewidth}
    \centering
    \includegraphics[width=0.8\linewidth]{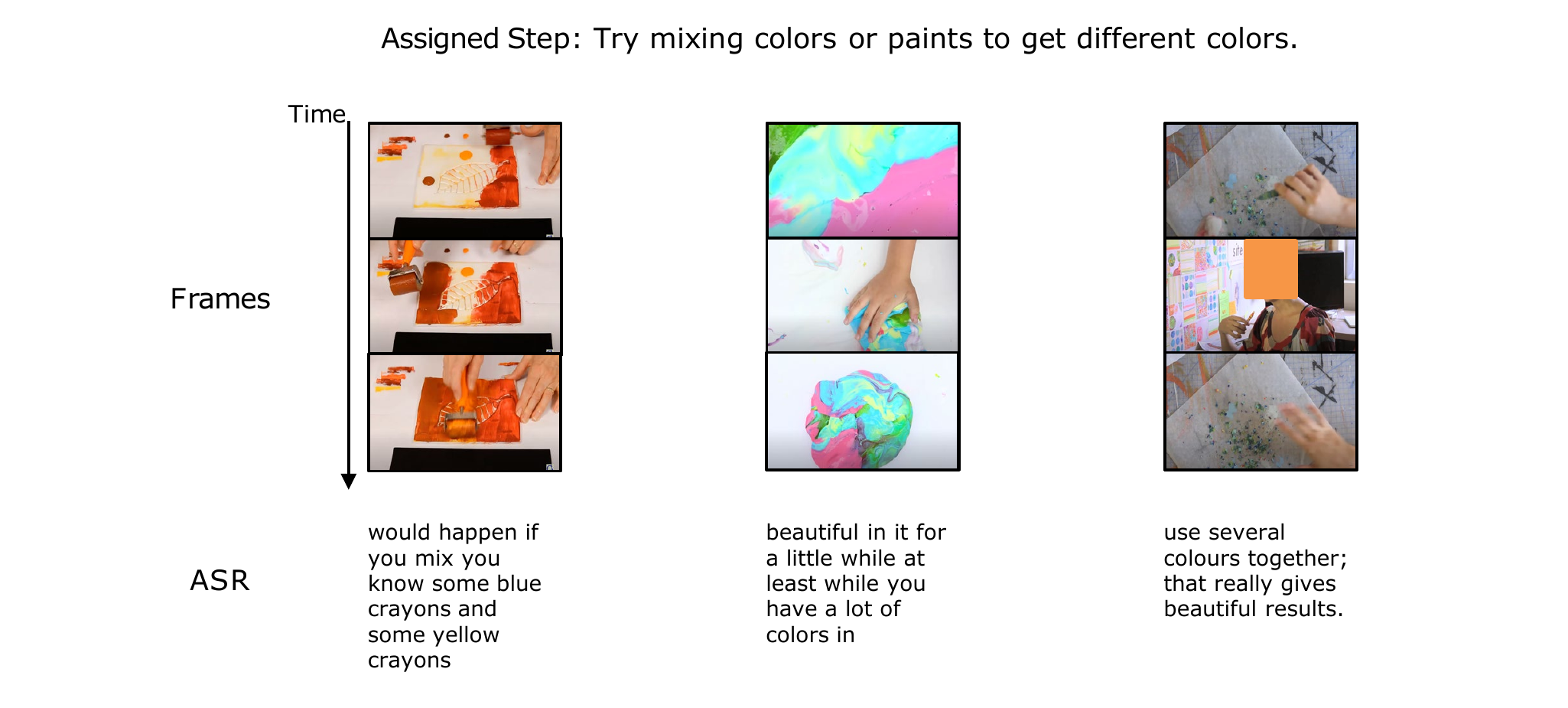}
    \caption{}
    \label{vis2:b}
\end{subfigure}
\begin{subfigure}[t]{\linewidth}
    \centering
    \includegraphics[width=0.8\linewidth]{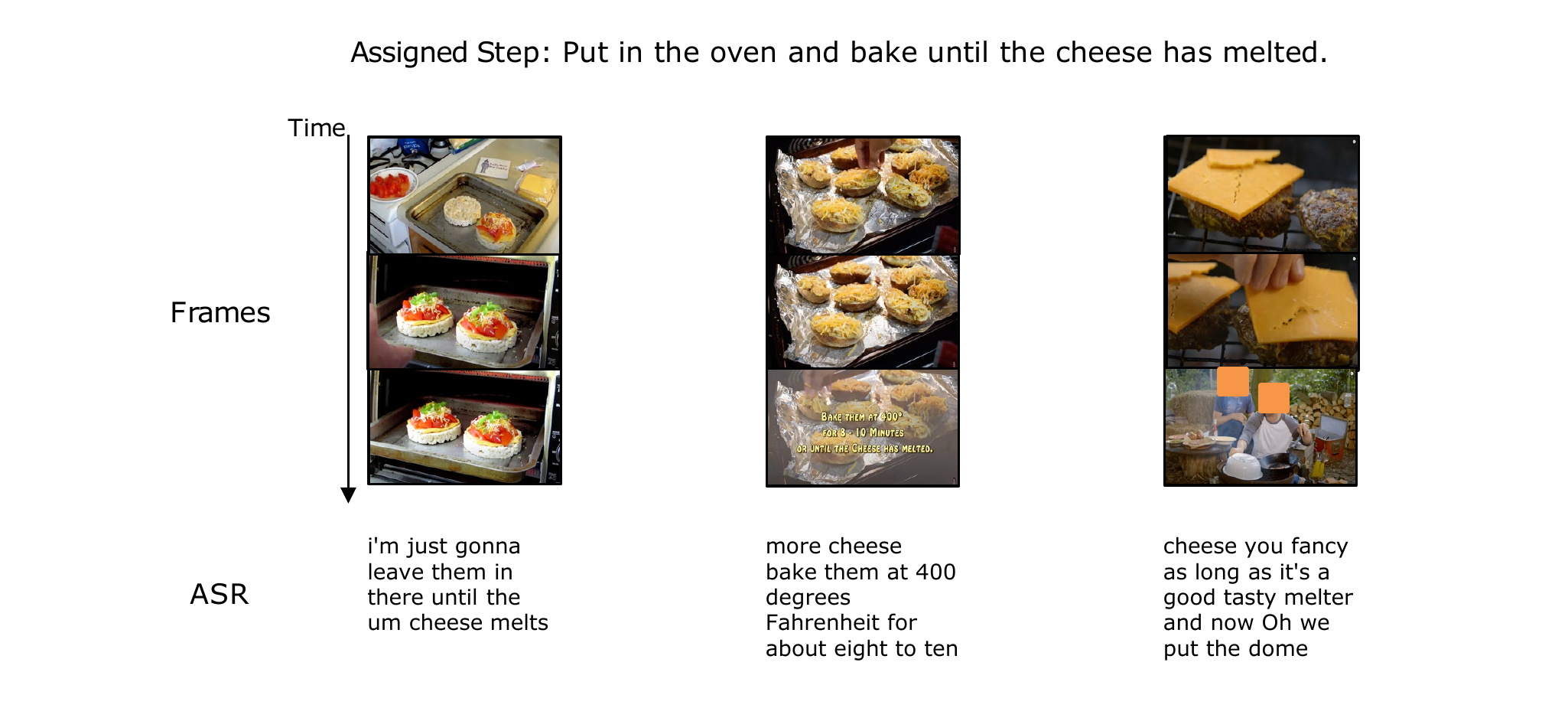}
    \caption{}
    \label{vis2:c}
\end{subfigure}
\caption{Visualization of segments assigned to three given steps by our distant supervision framework. We can observe that our method successfully groups together segments that are semantically coherent under each step, despite their large differences in appearance. Faces in the frames are artificially masked for privacy reasons.
}
\label{vis2}
\end{figure*}

\noindent\textbf{Visualization of Distant Supervision.} In Figure~\ref{vis1} we provide visualizations of steps assigned by our distant supervision method for three video examples. We can observe that the matched step descriptions capture high-level semantics about actions and objects, which are conversely often missed by the narrations. An example is given in Figure~\ref{vis1:a} where the narration in the last segment (``really really hot water you can use'') does not correspond to an object or an action directly recognizable in the segment. The language model assigns this narration to a more expressive step description (\textit{Open the hot water faucet in your sink or tub}). Figure~\ref{vis1:b} shows that the assigned steps capture higher level information compared to traditional atomic actions. For example, a video segment of pouring oil into a heated pan is matched to \textit{Prepare to fry tortillas}.  

\noindent\textbf{Visualization of Step Classes.} In order to better understand the variety of video segments that are grouped under a given step, in Figure~\ref{vis2} we show three video clips assigned to three given steps. We can observe that our method can successfully group together video segments that are coherent in terms of the demonstrated step. Note that, at the same time, the segments assigned to a given step exhibit large variations in terms of appearance (e.g., color, viewpoint, object instances). Because our model assigns segments to step descriptions purely based on language information, it is insensitive to these large appearance variations. This invariance is then transferred to the video model: by using these distantly supervised step classes as supervision for the video model, our method trains the video representation to be invariant to these appearance variations and to capture the higher-level semantics represented in each step class.


\begin{figure}[h]
\begin{center}
\includegraphics[width=0.9\linewidth]{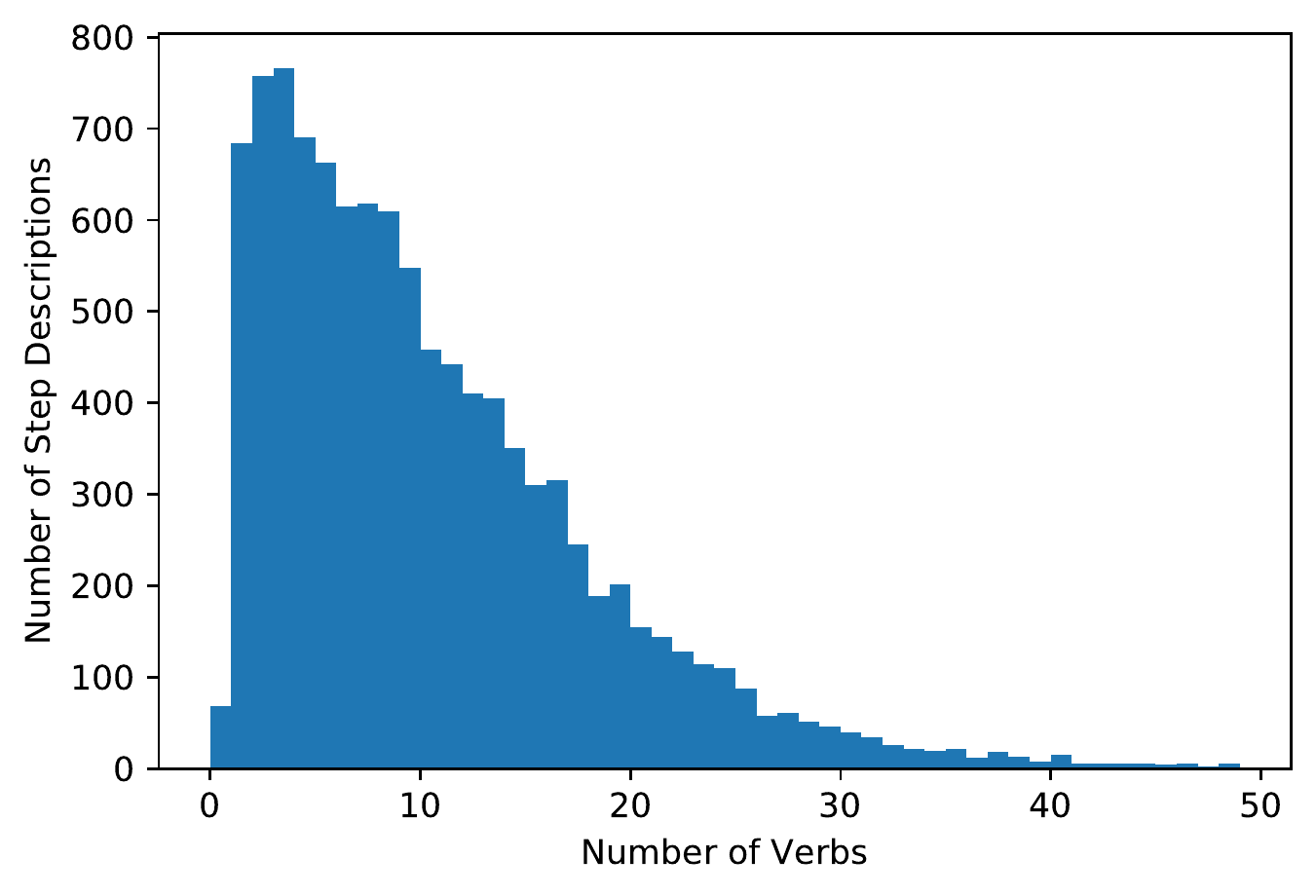}\vspace{-.5cm}
\end{center}
\caption{Histogram of number of verbs in wikiHow step descriptions.
}
\label{fig:step_stat}
\end{figure}

\noindent\textbf{Limitations and Failure Cases.}
Our approach may fail in assigning the correct step to a segment due to errors caused by the language model and due to excessive noise or ambiguity in the ASR sentence. Furthermore, the step description may refer to actions or objects not represented in the segment. For example, in Figure~\ref{vis2:c} the assigned step \textit{Put in the oven and bake until the cheese is melted} provides an accurate semantic description for the segments, but it refers to an object (``oven'') that is not shown in the video frames. On one hand, this visual misalignment may render the training difficult; on the other hand, it may still be beneficial, since it forces the model to use contextual information (e.g., visible objects that tend to co-occur with ``oven'', such as the bakeware objects appearing in the frames) to recognize the high-level semantics of the steps. Another potential limitation is the temporal misalignment between speech and visual content. However, this problem can be reduced by expanding the temporal span of the ASR text to increase the probability of including the relevant text for the given video segment, or by adopting a multiple-instance learning scheme~\cite{miech2020end} to find the correct temporal alignment between ASR text sentences and video segments.

\noindent\textbf{Complexity of Steps.}
In our experiments we demonstrated that, on the downstream problems of step and task classification, our distantly-supervised video representation outperforms video descriptors trained with full supervision on traditional action classes. We hypothesize that this is due to the fact that each step typically consists of multiple  actions performed in sequence, unlike traditional action classes which typically encode a single atomic action (e.g., ``drinking'', ``jumping'', ``punching''). To assess this hypothesis we analyzed the number of verbs returned by the POS tagger~\cite{nltk} for each wikiHow step description as a measure of the complexity of the step. Figure~\ref{fig:step_stat} shows the distribution of the number of verbs. The average and the median number of verbs in step descriptions are 10.1 and 8.0, respectively. Furthermore, more than $85\%$ of the steps contain at least 2 verbs. This indeed suggests that steps tend to have a higher-level of complexity compared to traditional atomic actions.

\begin{figure}[h]
\begin{center}
\includegraphics[width=0.9\linewidth]{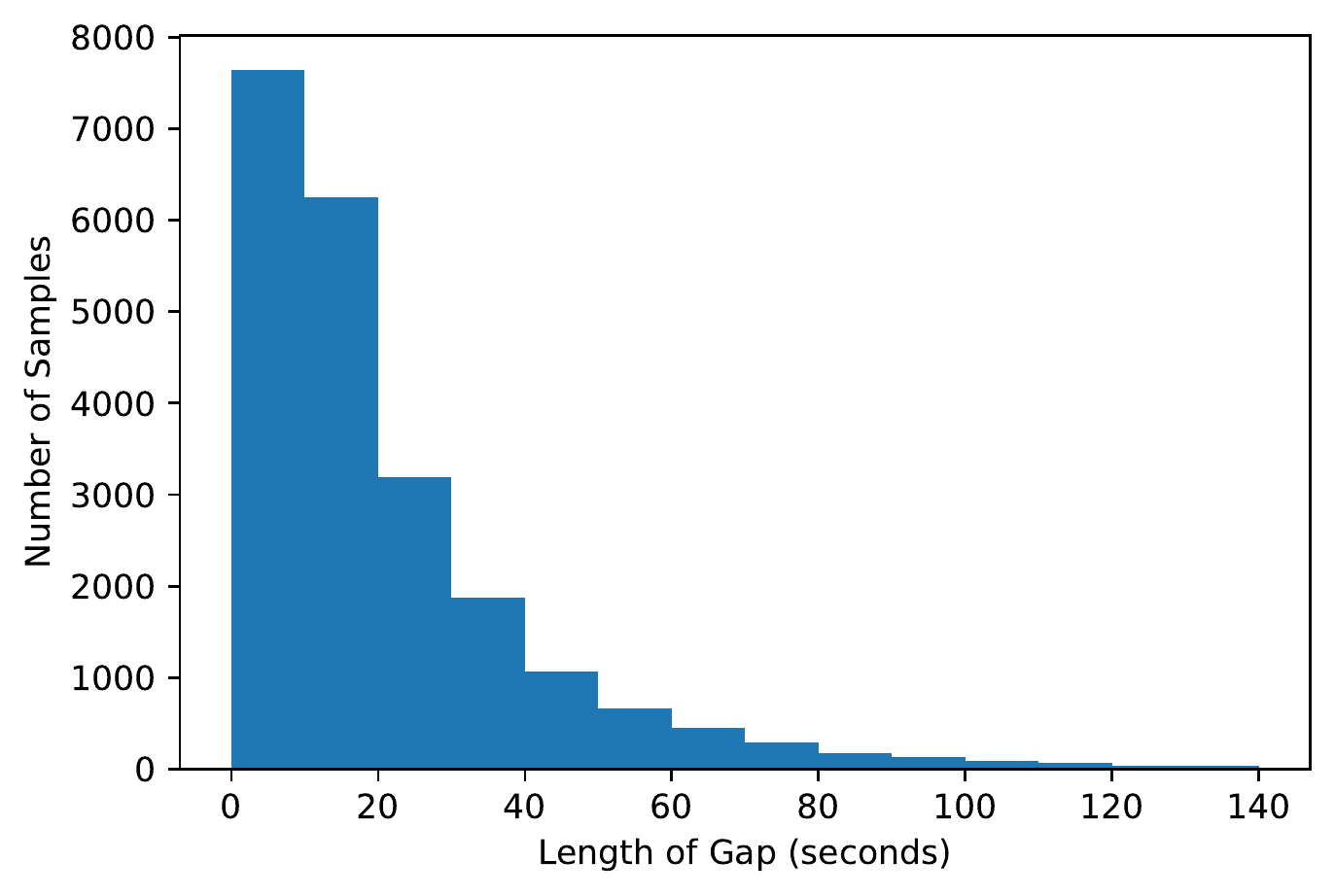}\vspace{-.5cm}
\end{center}
\caption{Histogram of the temporal gaps separating the observed history from the step to forecast in the COIN dataset. 
}
\label{fig:forecast_stat}
\end{figure}



\section{Further Details about Step Forecasting} 
We follow the training/validation split in the COIN dataset to train and evaluate our models for step forecasting. By constraining the observed history to contain at least one step, we construct a training set of 22037 samples and a validation set of 6721 samples.
Fig.~\ref{fig:forecast_stat} shows the distribution of the gaps (in seconds) separating the history from the step to predict. The average and the median of the gap are 21 seconds and 14 seconds, respectively. Thus, the forecasting gaps in this benchmarks are substantially longer than those used in other action anticipation tasks~\cite{ryoo2011human,hoai2014max,Gammulle_2019_ICCV}. This makes this benchmark particularly challenging as the model is asked to predict the step of segments far away in the future compared to the observed history.

\end{document}